\documentclass[10pt,twocolumn,letterpaper]{article}

\usepackage[pagenumbers]{cvpr} 

\usepackage{booktabs,multirow,bbding}

\definecolor{cvprblue}{rgb}{0.21,0.49,0.74}
\usepackage[pagebackref,breaklinks,colorlinks,allcolors=cvprblue]{hyperref}
\usepackage{xcolor}
\usepackage[table]{xcolor} 
\definecolor{mygrey}{cmyk}{0, 0, 0, 0.5}
\newcommand{\smallgrey}[1]{\textcolor{mygrey}{\scalebox{0.8}{#1}}}

\def\confName{CVPR}
\def\confYear{2027}

\title{Which Parameters to Update? An Instance-Specific Adaptation Framework for Real-World Pansharpening}

\author{Jin-Liang Xiao$^{1}$, Liang-Jian Deng$^{2}$\thanks{Corresponding author}, Yu-Xuan Zhang$^{1}$, Hua Meng$^{1}$, Zhengchun Zhou$^{1}$\\
$^{1}$ Southwest Jiaotong University, Chengdu, China\\
$^{2}$ University of Electronic Science and Technology of China, Chengdu, China\\
}

\begin{document}
\maketitle
\begin{abstract}
Pansharpening aims to fuse a single observed image pair, comprising a low-resolution multispectral (LRMS) image and a panchromatic (PAN) image, to obtain a high-resolution multispectral (HRMS) image. Recently, deep learning-based methods have achieved significant performance due to their strong capability of extracting nonlinear features. However, distribution shifts between training data and test images can degrade the performance of pretrained pansharpening networks, limiting their generalization in real-world scenarios. In this work, we propose an instance-specific test-time adaptation framework that uses the observed PAN–LRMS pair to select and update only a subset of the pretrained network’s parameters. Specifically, the framework uses cross-scoring to identify parameter groups whose updates improve reconstruction across image patches. It then updates the selected groups by minimizing a loss that combines a spectral consistency term with a newly designed spatial consistency term. Experiments show that the adapted networks perform better than their pretrained counterparts and outperform representative state-of-the-art pansharpening methods on GaoFen-2, QuickBird, and WorldView series datasets.
\end{abstract}    
\section{Introduction}
\label{sec:intro}
Pansharpening plays an important role in satellite-based Earth observation and other remote sensing applications by enhancing the spatial resolution of multispectral imagery \cite{yao2020cross, huang2024mftn, yu2024unmixing}. It fuses a low-resolution multispectral (LRMS) image with a high-resolution panchromatic (PAN) image to obtain a high-resolution multispectral (HRMS) image \cite{loncan2015hyperspectral}.

Pansharpening methods are commonly grouped into three categories, namely conventional fusion approaches, variational optimization (VO)-based methods, and deep learning (DL)-based methods \cite{vivone2024deep}. Conventional approaches typically rely on computationally efficient, closed-form fusion schemes based on component substitution or multiresolution analysis. VO-based methods formulate HRMS reconstruction as an optimization problem with spectral and spatial fidelity terms. Both families reconstruct an HRMS image from a single PAN–LRMS pair. Their reliance on predefined fusion rules or observation models limits their ability to capture complex nonlinear relationships between inputs and the target HRMS image \cite{deng2022machine}.

\begin{figure}[t]
    \centering
    \includegraphics[width=1\linewidth]{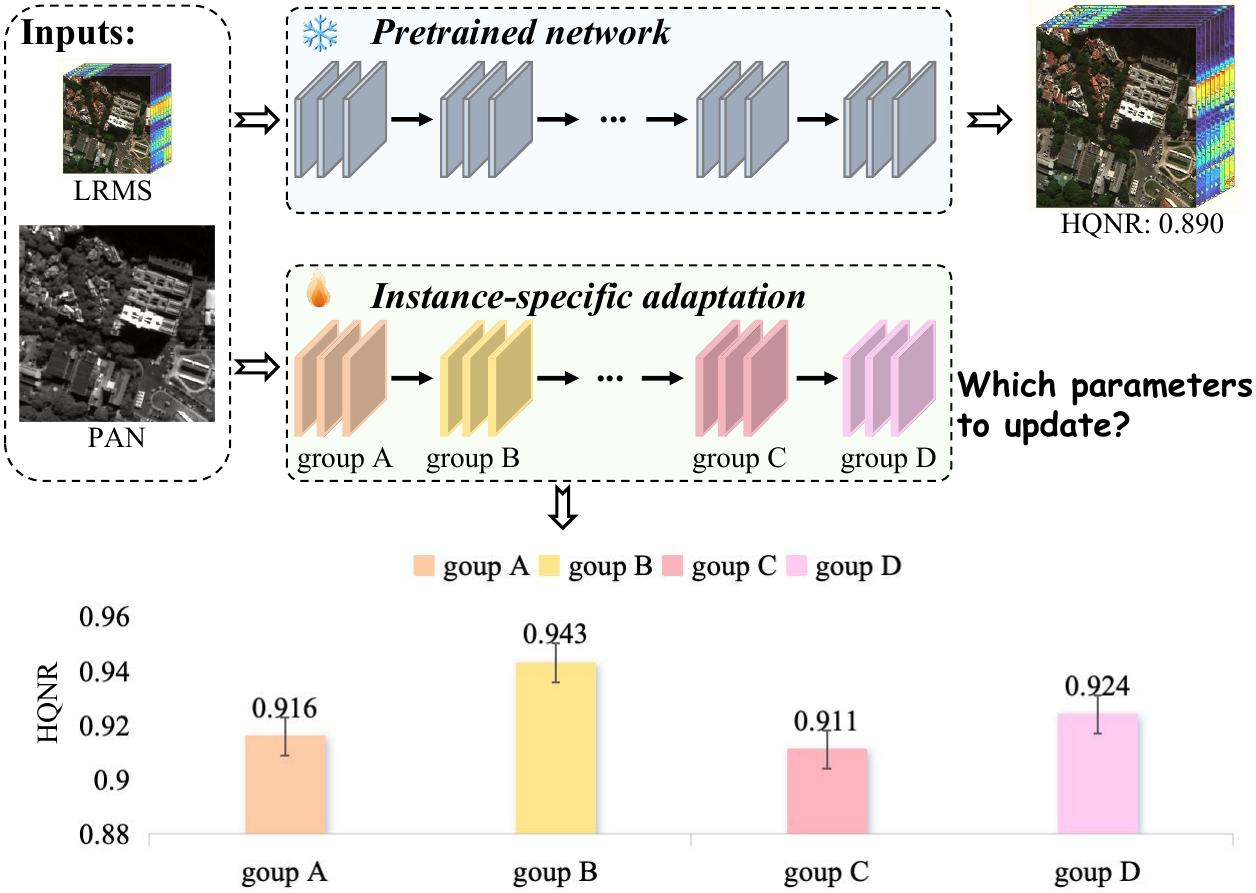}
    \caption{\textbf{Motivation for instance-specific parameter selection.} Given a test PAN–LRMS image pair, the pretrained network produces a limited HRMS image with an HQNR of 0.890. We evaluate the reconstruction performance obtained by adapting different parameter groups individually. The resulting HQNR values vary across groups, indicating that different parameter groups have unequal effects on the quality of the current test image. This observation motivates the selective adaptation of the most beneficial parameters rather than indiscriminately updating the entire network.}
    \label{fig:top}
\end{figure}

DL-based methods address this limitation by learning nonlinear mappings from PAN–LRMS pairs to HRMS images. Pansharpening methods based on convolutional neural networks (CNNs) \cite{deng2021fusionnet, meng2024progressive, xu2025laboring}, transformers \cite{hou2024linearly, su2023stcp}, diffusion models \cite{zhong2024ssdiff, meng2025accelerated, wang2025learning}, and Mamba \cite{peng2024fusionmamba, wang2026mmmamba} have improved reconstruction quality. Training these models typically requires large datasets and substantial computational resources. The resulting models may suffer performance degradation when test images differ in distribution from the training data \cite{xiao2025hyperspectral}. This challenge is particularly pronounced in real-world pansharpening, where image characteristics vary across sensors and acquisition conditions, while training data may cover only part of this variability.

Recently, instance-specific methods have attracted considerable attention because they can operate using only a single observed image pair, which can substantially reduce training costs. These methods typically either train the entire network on a small dataset synthesized from the observed pair or insert adaptation modules into intermediate layers of a pretrained network. This adaptation is challenging because it must draw on pretrained knowledge while keeping computational overhead low and relying only on the observed pair. However, existing methods often overlook differences in how parameters contribute to reconstruction for a given test pair. As shown in \Cref{fig:top}, parameters in different layers can contribute differently to HRMS reconstruction for the same pair. Updating the entire network or adding extra modules may therefore incur unnecessary computational overhead, and both strategies offer limited guidance on which pretrained parameters to adapt. \emph{This motivates two questions: (1) How can we identify the pretrained parameters most useful for adaptation to a given test pair? (2) How can we effectively update the selected parameters using only the observed pair?}

\begin{figure*}[t]
    \centering
    \includegraphics[width=1\linewidth]{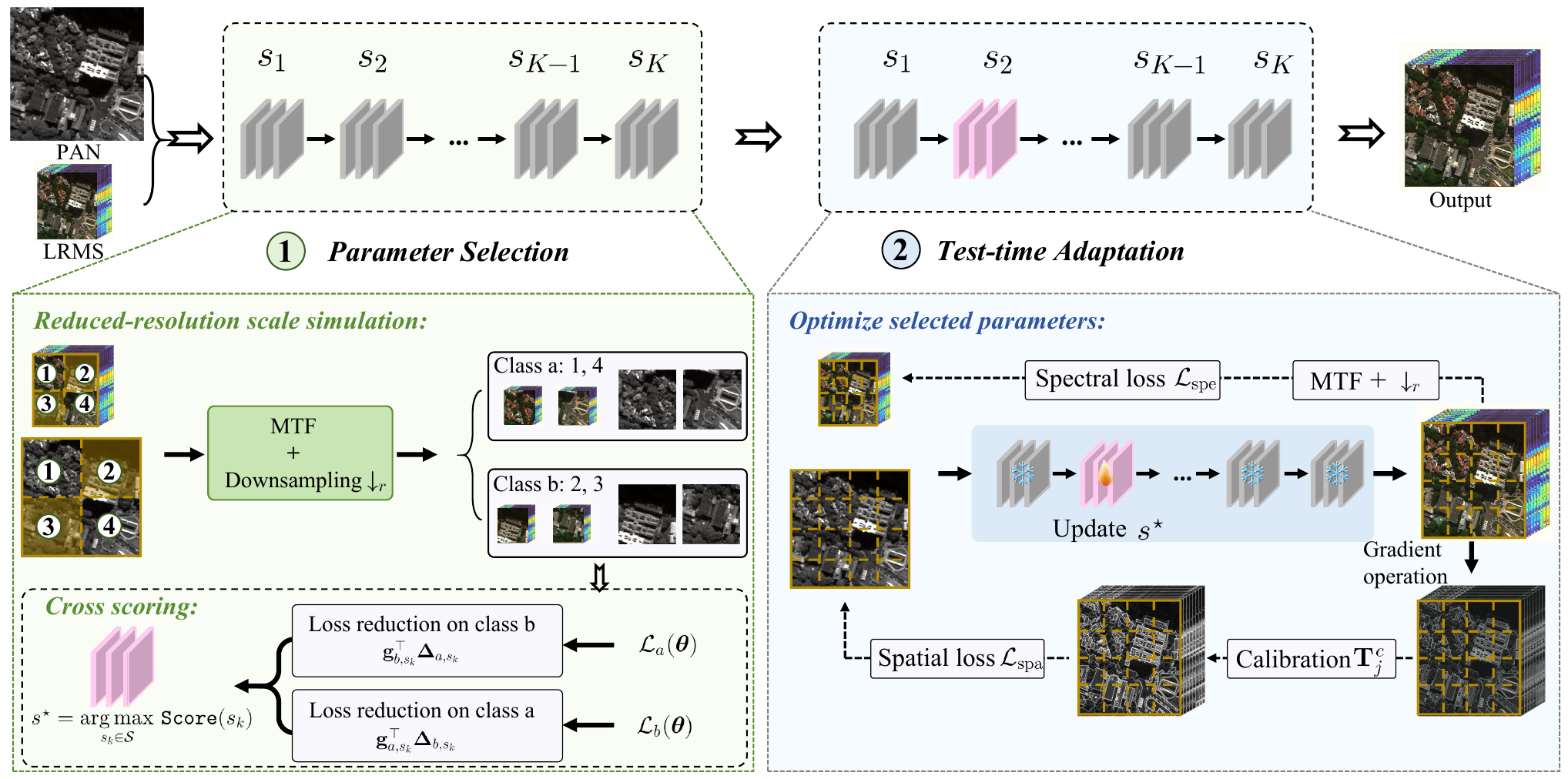}
    \caption{Illustration of the whole process of the proposed instance-specific adaptation framework that includes parameter selection and test-time adaptation. }
    \label{fig:framework}
\end{figure*}

We propose an instance-specific adaptation framework for pansharpening that updates only a selected group of pretrained parameters for each observed PAN--LRMS pair. As shown in \Cref{fig:framework}, the framework comprises two stages: parameter selection via cross-scoring and test-time adaptation.
\emph{In the parameter selection stage}, following Wald's protocol \cite{wald1997protocol}, we generate reduced-resolution samples from the observed PAN--LRMS pair and partition them into two spatially disjoint sets. We evaluate the candidate parameter groups under the same parameter budget. For each group, we compute a virtual first-step Adam update using gradients from one set. We then use a first-order approximation to estimate the resulting reduction in reconstruction loss on the other set. Averaging the estimated loss reductions in both directions yields a cross-score that measures how well the updates transfer across image regions. We select the group with the highest cross-score for adaptation. \emph{In the test-time adaptation stage}, we update only the selected parameter group using the original PAN--LRMS pair. The adaptation objective combines two complementary consistency losses. The spectral consistency loss matches the degraded HRMS reconstruction to the observed LRMS image. The spatial consistency loss uses PAN image gradients whose magnitudes are calibrated for each spectral band using the cached prediction from the pretrained network. We minimize the combined loss over the selected parameters and use early stopping to terminate adaptation.

The \emph{main contributions} are summarized as follows:
\begin{itemize}
    \item We propose an instance-specific pansharpening framework that adapts a pretrained network to each test pair by updating only a selected group of parameters using spectral and designed spatial consistency losses.
    \item We introduce a cross-scoring strategy that selects parameter groups by estimating how well their updates transfer across image regions within a single observed PAN--LRMS pair.
    \item Experiments on GaoFen-2, QuickBird, and WorldView series datasets show that the framework improves the performance of pretrained pansharpening networks and outperforms representative state-of-the-art methods.
\end{itemize}

\section{Related Work}

\subsection{Deep Learning-based Methods}

DL-based pansharpening methods learn nonlinear mappings from PAN and LRMS observations to the target HRMS image \cite{jiang2026observation, sun2026}. A central design choice in CNN-based pansharpening is how to incorporate fine spatial details from the PAN image into the multispectral reconstruction. PNN \cite{masi2016pnn} provided an early demonstration that CNNs could learn to fuse PAN and LRMS images directly from data. PanNet \cite{yang2017pannet} introduced high-pass-domain learning to recover fine spatial detail while preserving the spectral content of the LRMS image. DiCNN \cite{he2019dicnn} and FusionNet \cite{deng2021fusionnet} used residual learning and detail injection to predict spatial details that complement the upsampled LRMS image. ARConv \cite{Wang2025CVPR} adapts the receptive field to structures at different scales by learning the dimensions of rectangular kernels and adjusting the number of sampling points.

Other studies have explored how to transfer pansharpening models across resolutions and how to improve feature fusion \cite{xu2025hipandas}. STP-SOM \cite{zhang2023stp} uses an estimated spectral observation model to link learning at reduced and full resolutions. Wang \etal~\cite{wang2026mmmamba} proposed MMMamba, which uses a Mamba backbone with interleaved scanning to enable in-context fusion of PAN and LRMS features. GGF-PnP \cite{liang2026ggf} integrates gradient-domain guidance into existing pansharpening networks. However, DL-based methods typically require substantial training data, and their reconstruction quality can degrade when test images come from sensors or acquisition conditions that are underrepresented in the training data.

\subsection{Instance-Specific Methods}

Instance-specific learning exploits the complementary spectral and spatial information in a single observed image pair for high-resolution reconstruction. In hyperspectral sharpening, Dian \etal~\cite{dian2023zero} generated training samples by downsampling the observed hyperspectral and multispectral images. For pansharpening, Cao \etal~\cite{cao2024zero} introduced ZS-Pan, a plug-and-play framework that trains at both reduced and full resolutions using a single PAN--LRMS pair. Rui \etal~\cite{rui2024variational} embedded a neural network in a variational objective and jointly optimized the network parameters and the HRMS image. Optimizing a network separately for each test pair can be computationally expensive. To reduce this cost, ERFT \cite{xin2026training} trains a lightweight plug-in module that adjusts intermediate features of a pretrained network. The approaches discussed above specify the trainable components in advance, leaving open how to select a subset of pretrained parameters based on the observed test pair.

\section{Proposed Adaptation Framework}
\label{sec:method}

In this section, we describe the proposed instance-specific adaptation framework in detail. As illustrated in \Cref{fig:framework}, the framework consists of two stages: parameter selection and test-time adaptation. The first stage selects a group of pretrained parameters via cross-scoring, while the second stage adapts the selected parameters using the observed PAN--LRMS pair.

\subsection{Parameter Selection via Cross-Scoring}
\label{sec:parameter_selection}

\paragraph{Parameter candidates:}

For a pretrained network, we define a candidate pool
\(\mathcal{S}=\{s_1,\ldots,s_k,\ldots,s_K\}\), where each \(s_k\) denotes a group of scalar parameters from pretrained weight tensors. All candidates satisfy \(|s_k|=B\), where \(B\) is the parameter budget for adaptation. This fixed budget enables a fair comparison of the candidate parameter groups for the current test pair.

\paragraph{Reduced-resolution scale simulation:}
Subsequently, we select the candidate that is most suitable for the current test dataset. Since the ground-truth HRMS image is unavailable at test time, we construct a reduced-resolution surrogate task following Wald's protocol. This enables the evaluation of parameter candidates using the observed LRMS image as a reference. Although the surrogate loss differs from the full-resolution adaptation objective, it provides an observation-based signal for identifying parameter groups with transferable reconstruction capabilities. The current PAN--LRMS images are denoted as $\mathbf{P}$ and $\mathbf{Y}$, respectively. We partition the aligned observations into four corresponding patch pairs, denoted by $(\mathbf{P}_{i}, \mathbf{Y}_{i}), i = 1,\ldots,4$. For each patch pair $(\mathbf{P}_{i}, \mathbf{Y}_{i})$, we apply
modality-specific modulation transfer function (MTF) filtering \cite{aiazzi2006mtf} and
downsampling following Wald's protocol \cite{wald1997protocol}. The degraded patches
\(\widetilde{\mathbf{P}}_i\) and \(\widetilde{\mathbf{Y}}_i\) are used as network inputs,
and the observed LRMS patch \(\mathbf{Y}_{i}\) serves as the reconstruction
reference at the reduced-resolution scale. We place diagonally opposite patches in the same class, forming two classes, \(a\) and
\(b\). Thus, the reconstruction loss on class \(a\) is
\begin{equation}
    \mathcal{L}_{a}
    (\boldsymbol{\theta})
    =
    \frac{1}{2}
    \sum_{i\in\{1,4\}}
    \mathtt{MSE}
    \!\left(
        f_{\boldsymbol{\theta}}
        (\widetilde{\mathbf{P}}_i,\widetilde{\mathbf{Y}}_i),
        \mathbf{Y}_i
    \right),
    \label{eq:a_loss}
\end{equation}
where $\boldsymbol{\theta}$ denotes the pretrained parameters, and $\mathtt{MSE}$ is the mean squared error between the patch prediction and the patch reference. Similarly, the reconstruction loss on class \(b\), i.e., $\mathcal{L}_{b}
    (\boldsymbol{\theta})$, can be obtained.

\paragraph{Cross-scoring:}
The proposed cross-scoring strategy evaluates the transferability of parameter updates across different image regions by estimating whether updates derived from one patch set can also reduce the reconstruction loss on another. A parameter group that reduces the reconstruction loss on one region may not necessarily benefit other regions of the same test image. This motivates evaluating parameter candidates according to their cross-region transferability, rather than solely relying on their local gradient magnitudes. By assessing the estimated loss reduction across different regions, the proposed cross-scoring strategy favors parameter groups whose updates are more likely to provide consistent improvements within the current test PAN-LRMS image pair.

We evaluate the gradient of each class loss at the pretrained parameters
over the union of all candidate parameters. Let
\(\mathbf{g}_{a,s_k}\) denote the restriction of the resulting gradient
to candidate \(s_k\). The virtual update on class \(a\) is the first Adam
step from zero optimizer moments, restricted to \(s_k\). With coupled
weight decay \(\rho\), this step is
\begin{equation}
    \begin{aligned}
    \mathbf{h}_{a,s_k}
        &= \mathbf{g}_{a,s_k}
           + \rho\,\boldsymbol{\theta}_{s_k}, \\
    \boldsymbol{\Delta}_{a,s_k}
        &= -
           \eta \frac{\mathbf{h}_{a,s_k}}
                {|\mathbf{h}_{a,s_k}|+\epsilon},
    \end{aligned}
    \label{eq:virtual_adam_step}
\end{equation}
where the division is elementwise, $\eta$ is the learning rate, and \(\epsilon\) is the stabilization constant. Since all candidates share the same virtual learning rate, the common scaling factor can be omitted without affecting their ranking under the first-order approximation. To assess the reduction in the loss on class b after the virtual update on class a, we express the loss using a first-order Taylor expansion as follows:
\begin{equation}
    \mathcal{L}_{b}(\boldsymbol{\theta}+\boldsymbol{\Delta}_{a})
    =
    \mathcal{L}_{b}(\boldsymbol{\theta})
    +
    \mathbf{g}_{b,s_k}^{\top}\boldsymbol{\Delta}_{a,s_k} + \boldsymbol{O}( \left\|\boldsymbol{\Delta}_{a,s_k}\right\|^2),
    \label{eq:first_order_approximation}
\end{equation}
where $\cdot^{\top}$ is the transpose of a vector.
Therefore, for distinct classes \(a\) and \(b\), the cross reduction can be approximately represented as
\(-\mathbf{g}_{b,s_k}^{\top}\boldsymbol{\Delta}_{a,s_k}\). We average the estimates
in both directions as follows,
\begin{equation}
    \begin{aligned}
    \mathtt{Score}(s_k)
        &=
        -\frac{1}{2}
        \left(
            \mathbf{g}_{b,s_k}^{\top}\boldsymbol{\Delta}_{a,s_k}
            +
            \mathbf{g}_{a,s_k}^{\top}\boldsymbol{\Delta}_{b,s_k}
        \right).
    \end{aligned}
    \label{eq:bidirectional_cross_score}
\end{equation}
Each score is a sum of contributions from the weights in \(s_k\).
The two class gradients therefore suffice to score the entire candidate
pool, using one backward pass per class. The parameters in $s^\star$ are selected as 
\begin{equation}
    \begin{aligned}
    s^\star
        &= \underset{s_k \in \mathcal{S}}{\arg\max} ~\mathtt{Score}(s_k).
    \end{aligned}
    \label{eq:selected}
\end{equation}


\subsection{Test-Time Adaptation}
\label{sec:test_time_adaptation}

To adapt the selected parameter group \(s^\star\), we initialize the network with the pretrained parameters \(\boldsymbol{\theta}_0\) and make only the parameters in \(s^\star\) trainable. Instead of processing the entire image at once, we divide the aligned PAN--LRMS observations into corresponding patches of sizes \(h\times w\) and \(\frac{h}{r}\times\frac{w}{r}\), respectively, where \(r\) is the scale ratio. For each patch \(j\in\mathcal{M}\), let \(\overline{\mathbf{X}}_j=f_{\boldsymbol{\theta}}(\overline{\mathbf{P}}_j,\overline{\mathbf{Y}}_j)\) denote the network prediction. We cache the initial pretrained prediction, denoted by \(\mathbf{X}_j=f_{\boldsymbol{\theta}_0}(\overline{\mathbf{P}}_j,\overline{\mathbf{Y}}_j)\), before optimization begins.

\paragraph{Spectral consistency:}

In pansharpening, the PAN and LRMS observations are typically modeled as spectral and spatial degradations of the underlying HRMS image, respectively. We apply the image degradation model to each patch. For patch \(j\), the spatial degradation process can be expressed as
\begin{equation}
    \begin{aligned}
    \mathtt{MTF}(\overline{\mathbf{X}}_j)\downarrow_r = \overline{\mathbf{Y}}_j,
    \end{aligned}
    \label{eq:spe}
\end{equation}
where $\mathtt{MTF}$ denotes the blurring operation, and $\downarrow_r$ is the down-sampling operation with scale ratio $r$.
Therefore, the spectral loss is built as follows,
\begin{equation}
    \mathcal{L}_{\mathrm{spe}}
    =
    \frac{1}{|\mathcal{M}|}
    \sum_{j\in\mathcal{M}}
    \mathtt{MAE}
    \!\left(
        \mathtt{MTF}(\overline{\mathbf{X}}_j )\downarrow_r,
         \overline{\mathbf{Y}}_j
    \right),
    \label{eq:spectral_loss}
\end{equation}
where \(|\mathcal{M}|\) is the number of patches, and \(\mathtt{MAE}\)
averages the absolute error.

\paragraph{Spatial consistency:}

\begin{figure}[t]
    \centering
    \includegraphics[width=0.9\linewidth]{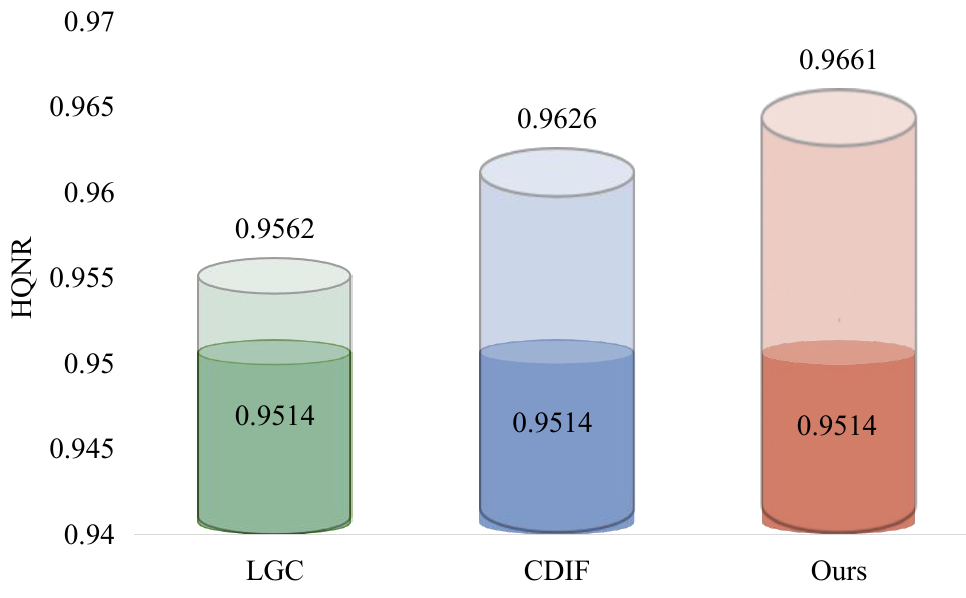}
    \caption{Effectiveness of the proposed spatial consistency term compared with two existing spatial fidelity terms, LGC \cite{fu2019variational} and CDIF \cite{xiao2022cdif} ($0.9514$ is the HQNR value using spectral fidelity). }
    \label{fig:spa}
\end{figure}

We use the PAN image to provide spatial detail guidance and the pretrained prediction to calibrate the gradient magnitude for each spectral band. Previous studies have demonstrated the effectiveness of gradient-domain constraints in reducing spatial discrepancies between PAN and HRMS images \cite{fu2019variational, xiao2022cdif}. Motivated by these findings, we explore the relationship between PAN and HRMS images in the gradient domain. Let \(\nabla\) denote the finite-difference operator. For band \(c\) in patch \(j\), the calibrated gradient target is defined as

\begin{equation}
    \mathbf{T}_{j}^{c}
    =
    \frac{
        \operatorname{std}
        \!\left(\nabla\mathbf{X}^{c}_{j}\right)
    }{
        \operatorname{std}
        \!\left(\nabla\overline{\mathbf{P}}_j\right)
        + \epsilon
    }\,
    \nabla\overline{\mathbf{P}}_j,
    \label{eq:pan_gradient_target}
\end{equation}
where the standard deviations are computed over spatial locations
within the patch. Directly imposing PAN gradients on individual MS bands may introduce excessive spatial variations due to differences in their spectral responses. To mitigate this issue, we calibrate the PAN gradient magnitude using the band-specific gradient statistics of the pretrained prediction. This allows the spatial consistency term to incorporate PAN-derived structural information while accounting for the spatial characteristics of each predicted spectral band. The spatial loss compares the predicted gradients with these fixed targets.
\begin{equation}
    \mathcal{L}_{\mathrm{spa}}
    =
    \frac{
        \displaystyle
        \sum_{c}\sum_{\mathcal{M}}
        \mathtt{MSE}
        \!\left(\nabla\overline{\mathbf{X}}^{c}_{j},\mathbf{T}^{c}_{j}\right)
    }{
        \max\!\left\{
            \left[
                \mathtt{mean}_{\mathcal{M}}
                \!\left(\mathbf{X}\right)
            \right]^2,
            \epsilon
        \right\}
    },
    \label{eq:spatial_loss}
\end{equation}
where \(\mathtt{mean}_{\mathcal{M}}\) averages over patches, spectral
bands, and spatial locations. The spatial loss is normalized by the squared mean intensity of the pretrained predictions to reduce its sensitivity to radiometric scaling across different scenes. As shown in \Cref{fig:spa}, the proposed spatial consistency term yields better reconstruction performance than the two compared spatial fidelity terms.


\paragraph{Optimization and Stopping Criterion:}
Combining the spectral consistency loss in
\cref{eq:spectral_loss} and the spatial consistency loss in
\cref{eq:spatial_loss}, the overall adaptation objective is
formulated as
\begin{equation}
    \mathcal{L}
    =
    \mathcal{L}_{\mathrm{spe}}
    +
    \lambda
    \mathcal{L}_{\mathrm{spa}},
    \label{eq:joint_adaptation_loss}
\end{equation}
where $\lambda$ is a trade-off parameter
balancing spectral and spatial consistency.
During adaptation, only the selected parameters in
$s^\star$ are updated, while the remaining pretrained
parameters are kept fixed.

To prevent overfitting to the current image pair, we reserve two patches for validation, which are excluded from parameter updates. The adaptation is performed for at most $N$ iterations. We evaluate the same objective on the held-out patches at initialization and every five iterations thereafter.
Early stopping is triggered if the validation loss fails to improve for three consecutive evaluations.

\begin{figure*}[t]
    \centering
    \includegraphics[width=1\linewidth]{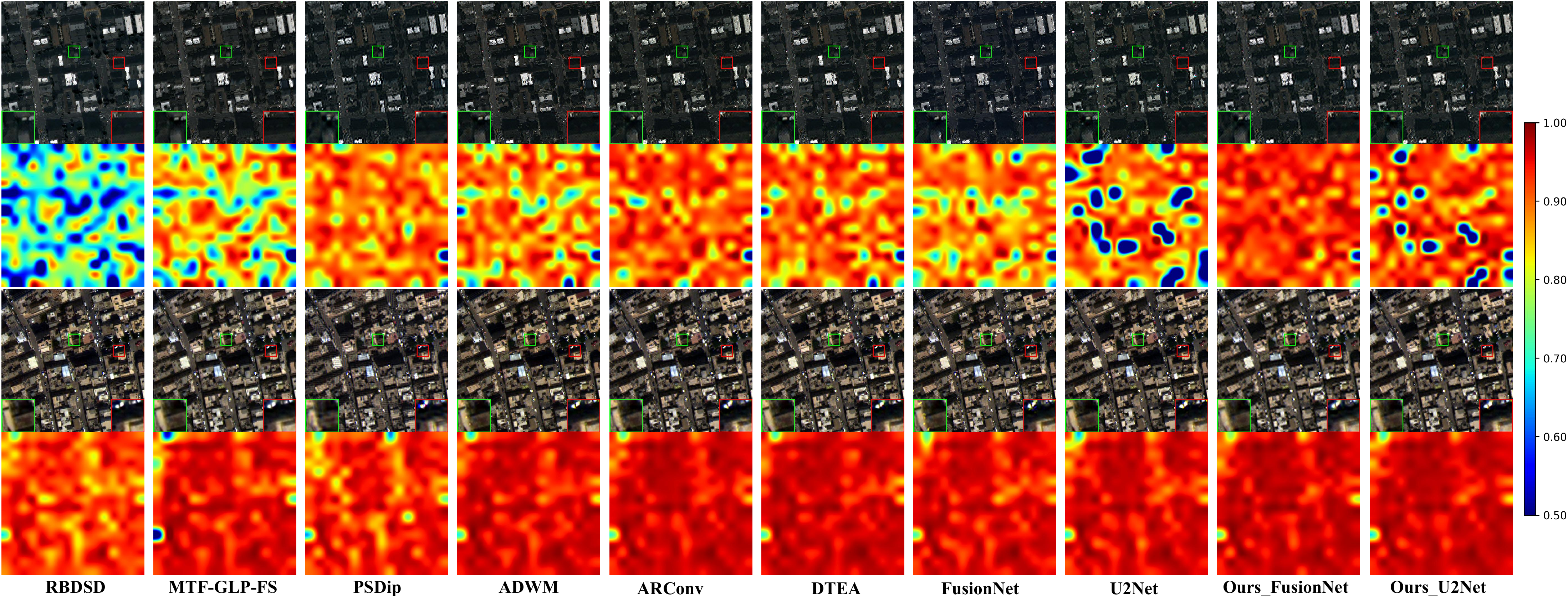}
    \caption{The in-sensor visual comparisons on QB and WV-3 datasets. The upper row shows the predicted HRMS images of different methods, and the bottom row displays the corresponding HQNR maps.  }
    \label{fig:insensor}
\end{figure*}

\begin{table*}[htbp!]
  \centering
  \footnotesize
  \setlength{\tabcolsep}{1mm}
  \renewcommand{\arraystretch}{1.1}
  \caption{Full-resolution quantitative in-sensor comparison on WV-3 and QB datasets. Results are reported as mean $\pm$ standard
  deviation over 20 image pairs per dataset. (Bold: best; Underline: second best)}
  \label{tab:insensor}
  \resizebox{\textwidth}{!}{%
    \begin{tabular}{@{}l|ccc|ccc@{}}
      \hline\hline
      \multirow{2}{*}{Method}
        & \multicolumn{3}{c|}{WV-3}
        & \multicolumn{3}{c}{QB} \\
      \cline{2-7}
        & HQNR $\uparrow$ & $D_\lambda \downarrow$ & $D_s \downarrow$
        & HQNR $\uparrow$ & $D_\lambda \downarrow$ & $D_s \downarrow$ \\
      \hline
      RBDSD \smallgrey{(19'TGRS)} \cite{vivone2019robust}
        & $0.8673\pm0.0543$ & $0.0634\pm0.0246$ & $0.0749\pm0.0359$
        & $0.6643\pm0.0561$ & $0.2003\pm0.0329$ & $0.1704\pm0.0472$ \\
      MTF-GLP-FS \smallgrey{(18'TIP)} \cite{vivone2018full}
        & $0.9127\pm0.0348$ & $0.0357\pm0.0106$ & $0.0537\pm0.0273$
        & $0.7941\pm0.0356$ & $0.0695\pm0.0198$ & $0.1470\pm0.0223$ \\
      PSDip \smallgrey{(24'TGRS)} \cite{rui2024variational}
        & $0.9222\pm0.0172$ & $0.0192\pm0.0076$ & $0.0598\pm0.0112$
        & $0.9020\pm0.0129$ & $\underline{0.0315\pm0.0178}$ & $0.0684\pm0.0210$ \\
      ADWM \smallgrey{(25'CVPR)} \cite{huang2025general}
        & $0.9277\pm0.0286$ & $0.0306\pm0.0118$ & $0.0432\pm0.0194$
        & $0.8581\pm0.0313$ & $0.0770\pm0.0185$ & $0.0706\pm0.0172$ \\
      ARConv \smallgrey{(25'CVPR)} \cite{Wang2025CVPR}
        & $0.9579\pm0.0103$ & $0.0146\pm0.0059$ & $0.0279\pm0.0068$
        & $\underline{0.9236\pm0.0191}$ & $0.0384\pm0.0148$ & $\mathbf{0.0396\pm0.0090}$ \\
      DTEA \smallgrey{(26'TIP)} \cite{xu2026deta}
        & $0.9447\pm0.0227$ & $0.0195\pm0.0081$ & $0.0366\pm0.0168$
        & $0.9021\pm0.0262$ & $0.0608\pm0.0183$ & $\mathbf{0.0396\pm0.0112}$ \\

      FusionNet \smallgrey{(21'TGRS)} \cite{deng2021fusionnet}
        & $0.9414\pm0.0196$ & $0.0216\pm0.0079$ & $0.0378\pm0.0142$
        & $0.8772\pm0.0201$ & $0.0674\pm0.0196$ & $0.0594\pm0.0121$ \\
      U2Net \smallgrey{(23'MM)} \cite{peng2023u2net}
        & $0.9507\pm0.0114$ & $0.0174\pm0.0069$ & $0.0325\pm0.0073$
        & $0.8082\pm0.0279$ & $0.1182\pm0.0170$ & $0.0835\pm0.0225$ \\

      \rowcolor{gray!20}Ours$_{\mathrm{FusionNet}}$
        & $\mathbf{0.9661\pm0.0049}$ & $\mathbf{0.0079\pm0.0037}$ & $\underline{0.0263\pm0.0042}$
        & $\mathbf{0.9356\pm0.0300}$ & $\mathbf{0.0210\pm0.0112}$ & $0.0446\pm0.0215$ \\
      \rowcolor{gray!20}Ours$_{\mathrm{U2Net}}$
        & $\underline{0.9615\pm0.0115}$ & $\underline{0.0133\pm0.0055}$ & $\mathbf{0.0256\pm0.0079}$
        & $0.8977\pm0.0274$ & $0.0514\pm0.0194$ & $0.0537\pm0.0159$ \\
      \hline\hline
    \end{tabular}%
  }
\end{table*}

\section{Numerical Experiments}

In this section, we evaluate the proposed instance-specific adaptation framework on real-world pansharpening scenes under both in-sensor and cross-sensor settings. The in-sensor experiments evaluate the improvement in reconstruction quality when the test sensor is represented in the pretraining data. The cross-sensor experiments assess whether the proposed adaptation can also improve reconstruction quality for sensors unseen during pretraining. Ablation studies examine the contributions of cross-scoring-based parameter selection and the proposed spatial consistency loss to reconstruction quality. In addition, we provide further discussions on comparisons with existing instance-specific methods, computational efficiency, and limitations.

\subsection{Datasets, Benchmark, and Metrics}

\paragraph{Datasets:}
The experiments use the pansharpening datasets from PanCollection \cite{deng2022machine} that include GaoFen-2 (GF-2), QuickBird (QB), WorldView-2 (WV-2), and WorldView-3 (WV-3) data. Each dataset contains 20 images from its corresponding sensor, and no reference is available at this scale.


\paragraph{Benchmark:} To demonstrate the effectiveness of the proposed method, the compared methods include: RBDSD \smallgrey{(19'TGRS)} \cite{vivone2019robust}, MTF-GLP-FS \smallgrey{(18'TIP)} \cite{vivone2018full}, PSDip \smallgrey{(24'TGRS)} \cite{rui2024variational}, ADWM  \smallgrey{(25'CVPR)} \cite{huang2025general}, ARConv \smallgrey{(25'CVPR)} \cite{Wang2025CVPR}, DTEA \smallgrey{(26'TIP)} \cite{xu2026deta}, FusionNet \smallgrey{(21'TGRS)} \cite{deng2021fusionnet}, U2Net \smallgrey{(23'MM)} \cite{peng2023u2net}, and FusionNet and U2Net are also used as the pretrained backbones of the proposed framework.

\paragraph{Metrics:} Performance on real-world data is evaluated by spectral distortion $D_{\lambda}$, spatial distortion $D_{s}$, and the hybrid quality with no reference index 
\begin{equation}
    \operatorname{HQNR}=(1-D_{\lambda})(1-D_{s}),
\end{equation}
following no-reference pansharpening assessment \cite{arienzo2022full}.

\subsection{Results on In-Sensor Data}
In this part, we assess performance on in-sensor data, for which the training and test data come from the same sensor. \Cref{tab:insensor} reports the quantitative results of all compared methods on the QB and WV-3 full-resolution datasets. As shown in \Cref{tab:insensor}, the proposed framework consistently improves the performance of the adopted pretrained backbones and achieves the best overall results on both datasets. In particular, compared with the original backbone networks, the refined results exhibit lower spectral and spatial distortions, leading to improved HQNR scores. These results indicate that the proposed framework is effective under the challenging full-resolution setting, where no ground-truth fused image is available. \Cref{fig:insensor} further presents the visual comparisons among different methods, together with the corresponding HQNR maps. The proposed method produces visually sharper spatial structures while maintaining better spectral consistency.

\subsection{Results on Cross-Sensor Data}

\begin{figure*}[t]
    \centering
    \includegraphics[width=1\linewidth]{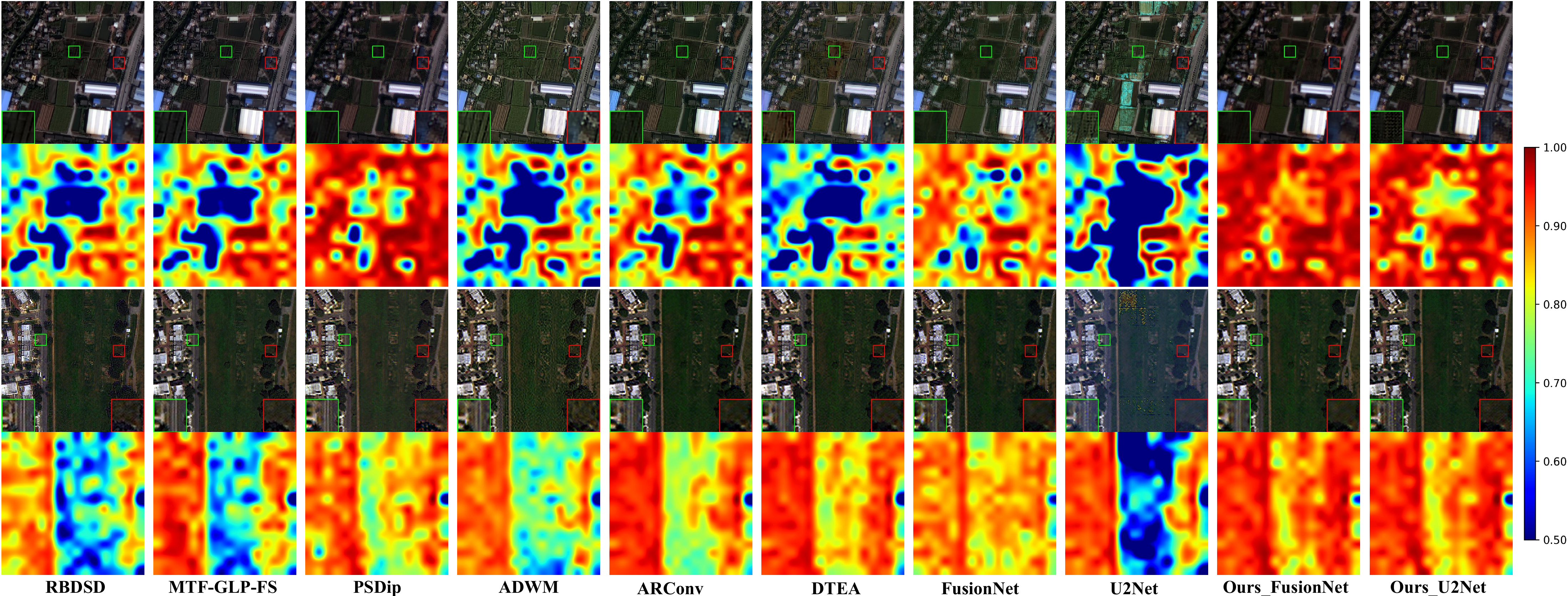}
    \caption{The cross-sensor visual comparisons on GF-2 and WV-2 datasets. The upper row shows the predicted HRMS images of different methods, and the bottom row displays the corresponding HQNR maps. }
    \label{fig:crosssensor}
\end{figure*}

\Cref{tab:gf2_wv2_comparison} evaluates the generalization ability of the proposed framework under cross-sensor shifts, where pretrained networks are applied to image pairs acquired by different sensors. Compared with the in-sensor setting, DL-based methods exhibit significant performance degradation. The proposed adaptation consistently improves both backbones over direct inference, demonstrating that its effectiveness is not limited to in-sensor scenarios. For QB $\rightarrow$ GF-2, $\mathrm{Ours}_{\mathrm{FusionNet}}$ achieves the best overall performance. Although $\mathrm{Ours}_{\mathrm{U2Net}}$ achieves a lower HQNR than PSDip \cite{rui2024variational}, it substantially outperforms the pretrained U2Net baseline \cite{peng2023u2net}. For WV-3 $\rightarrow$ WV-2, both adapted variants also achieve the best overall results. The visual comparisons in \Cref{fig:crosssensor} further support these findings. Compared with the pretrained backbones, our results preserve clearer building boundaries and finer spatial structures while achieving higher HQNR values in the corresponding maps.

\begin{table*}[htbp!]
  \centering
  \footnotesize
  \setlength{\tabcolsep}{1mm}
  \renewcommand{\arraystretch}{1.1}
  \caption{Quantitative cross-sensor comparison on GF-2 and WV-2 datasets, using pretrained networks trained on QB and WV-3, respectively. Results are reported as mean $\pm$ standard
  deviation over 20 image pairs per dataset. (Bold: best; Underline: second best)}
  \label{tab:gf2_wv2_comparison}
  \resizebox{\textwidth}{!}{%
    \begin{tabular}{@{}l|ccc|ccc@{}}
      \hline\hline
      \multirow{2}{*}{Method}
        & \multicolumn{3}{c|}{QB $\rightarrow$ GF2}
        & \multicolumn{3}{c}{WV-3$\rightarrow$ WV-2} \\
      \cline{2-7}
        & HQNR $\uparrow$ & $D_\lambda \downarrow$ & $D_s \downarrow$
        & HQNR $\uparrow$ & $D_\lambda \downarrow$ & $D_s \downarrow$ \\
      \hline
      RBDSD \smallgrey{(19'TGRS)} \cite{vivone2019robust}
        & $0.7717\pm0.0442$ & $0.0841\pm0.0336$ & $0.1578\pm0.0280$
        & $0.8286\pm0.0432$ & $0.1413\pm0.0320$ & $0.0356\pm0.0213$ \\
      MTF-GLP-FS \smallgrey{(18'TIP)} \cite{vivone2018full}
        & $0.8238\pm0.0371$ & $0.0500\pm0.0166$ & $0.1331\pm0.0291$
        & $0.8658\pm0.0415$ & $0.0563\pm0.0212$ & $0.0830\pm0.0260$ \\
      PSDip \smallgrey{(24'TGRS)} \cite{rui2024variational}
        & $\underline{0.9489\pm0.0290}$ & $\underline{0.0255\pm0.0216}$ & $\underline{0.0264\pm0.0137}$
        & $0.8935\pm0.0259$ & $0.0412\pm0.0276$ & $0.0680\pm0.0133$ \\
      ADWM \smallgrey{(25'CVPR)} \cite{huang2025general}
        & $0.7220\pm0.0413$ & $0.2173\pm0.0482$ & $0.0769\pm0.0261$
        & $0.8686\pm0.0411$ & $0.0490\pm0.0402$ & $0.0866\pm0.0205$ \\
      ARConv \smallgrey{(25'CVPR)} \cite{Wang2025CVPR}
        & $0.8626\pm0.0357$ & $0.0523\pm0.0306$ & $0.0898\pm0.0252$
        & $0.9254\pm0.0344$ & $0.0461\pm0.0318$ & $0.0299\pm0.0114$ \\
      DTEA \smallgrey{(26'TIP)} \cite{xu2026deta}
        & $0.7194\pm0.0580$ & $0.2422\pm0.0606$ & $0.0505\pm0.0155$
        & $0.9351\pm0.0281$ & $0.0452\pm0.0283$ & $\mathbf{0.0207\pm0.0062}$ \\
      FusionNet \smallgrey{(21'TGRS)} \cite{deng2021fusionnet}
        & $0.8241\pm0.0604$ & $0.1163\pm0.0549$ & $0.0681\pm0.0173$
        & $0.8906\pm0.0197$ & $0.0505\pm0.0249$ & $0.0618\pm0.0176$ \\
      U2Net \smallgrey{(23'MM)} \cite{peng2023u2net}
        & $0.7071\pm0.0717$ & $0.2230\pm0.0638$ & $0.0907\pm0.0443$
        & $0.8687\pm0.0842$ & $0.0950\pm0.0826$ & $0.0407\pm0.0100$ \\
      \rowcolor{gray!20}Ours$_{\mathrm{FusionNet}}$
        & $\mathbf{0.9577\pm0.0271}$ & $\mathbf{0.0217\pm0.0227}$ & $\mathbf{0.0212\pm0.0080}$
        & $\underline{0.9434\pm0.0252}$ & $\mathbf{0.0249\pm0.0162}$ & $0.0326\pm0.0133$ \\
      \rowcolor{gray!20}Ours$_{\mathrm{U2Net}}$
        & $0.9408\pm0.0264$ & $0.0268\pm0.0183$ & $0.0333\pm0.0144$
        & $\mathbf{0.9450\pm0.0263}$ & $\underline{0.0289\pm0.0174}$ & $\underline{0.0270\pm0.0126}$ \\
      \hline\hline
    \end{tabular}%
  }
\end{table*}

\begin{table}[htbp!]
  \centering
  \footnotesize
  \setlength{\tabcolsep}{1.0mm}
  \renewcommand{\arraystretch}{1}
  \caption{Comparison of parameter selection strategies on the WV-2 dataset, including different fixed selections and the proposed cross-scoring strategy. (Bold: best; underline: second best)}
  \label{tab:ablation_selection}
  \resizebox{\columnwidth}{!}{%
    \begin{tabular}{@{}l|l|ccc@{}}
      \toprule
      \textbf{Backbone} & \textbf{Selection strategy} & HQNR $\uparrow$ & $D_\lambda \downarrow$ & $D_s \downarrow$ \\
      \midrule
      \multirow{4}{*}{FusionNet}
       & Early selection & $0.9246\pm0.0227$ & $\underline{0.0359\pm0.0213}$ & $0.0410\pm0.0131$ \\
       & Middle selection & $\underline{0.9308\pm0.0245}$ & $0.0396\pm0.0233$ & $\mathbf{0.0308\pm0.0102}$ \\
       & Late selection & $0.9116\pm0.0243$ & $0.0407\pm0.0210$ & $0.0497\pm0.0161$ \\
       & Cross-scoring & $\mathbf{0.9434\pm0.0252}$ & $\mathbf{0.0249\pm0.0162}$ & $\underline{0.0326\pm0.0133}$ \\
      \midrule
      \multirow{4}{*}{U2Net}
       & Early selection & $0.9253\pm0.0305$ & $0.0373\pm0.0232$ & $0.0388\pm0.0109$ \\
       & Middle selection & $0.8700\pm0.0834$ & $0.0936\pm0.0818$ & $0.0407\pm0.0099$ \\
       & Late selection & $\underline{0.9419\pm0.0275}$ & $\underline{0.0303\pm0.0186}$ & $\underline{0.0288\pm0.0139}$ \\
       & Cross-scoring  & $\mathbf{0.9450\pm0.0263}$ & $\mathbf{0.0289\pm0.0174}$ & $\mathbf{0.0270\pm0.0126}$ \\
      \bottomrule
    \end{tabular}%
  }
\end{table}

\begin{table}[htbp!]
  \centering
  \footnotesize
  \setlength{\tabcolsep}{2.0mm}
  \renewcommand{\arraystretch}{1.1}
  \caption{Spatial and spectral loss ablation on 20 full-resolution WV-3 image pairs. Values are mean $\pm$ standard deviation across images. (Bold: best; Underline: second best) }
  \label{tab:ablation_losses}
  \resizebox{\columnwidth}{!}{%
    \begin{tabular}{@{}l|cc|ccc@{}}
      \toprule
      \textbf{Backbone} & $\mathcal{L}_{\mathrm{spa}}$ & $\mathcal{L}_{\mathrm{spe}}$ & HQNR $\uparrow$ & $D_\lambda \downarrow$ & $D_s \downarrow$ \\
      \midrule
      \multirow{3}{*}{FusionNet} 
       & \CheckmarkBold &  & $0.5000\pm0.1419$ & $0.3966\pm0.1429$ & $0.1798\pm0.0445$ \\
       &  & \CheckmarkBold & \underline{$0.9514\pm0.0153$} & $\mathbf{0.0075\pm0.0033}$ & \underline{$0.0414\pm0.0138$} \\
       & \CheckmarkBold & \CheckmarkBold & $\mathbf{0.9661\pm0.0049}$ & \underline{$0.0079\pm0.0037$} & $\mathbf{0.0263\pm0.0042}$ \\
      \midrule
      \multirow{3}{*}{U2Net} 
       & \CheckmarkBold &  & $0.6084\pm0.0616$ & $0.2881\pm0.0638$ & $0.1462\pm0.0219$ \\
       &  & \CheckmarkBold & $\underline{0.9429\pm0.0153}$ & $\mathbf{0.0128\pm0.0048}$ & $\underline{0.0449\pm0.0119}$ \\
       & \CheckmarkBold & \CheckmarkBold & $\mathbf{0.9615\pm0.0115}$ & $\underline{0.0133\pm0.0055}$ & $\mathbf{0.0256\pm0.0079}$ \\
      \bottomrule
    \end{tabular}%
  }
\end{table}

\begin{table*}[t]
  \centering
  \footnotesize
  \setlength{\tabcolsep}{1mm}
  \renewcommand{\arraystretch}{1.1}
  \caption{Full-resolution quality and runtime comparison on WV-3 and WV-2.
  All pretrained configurations use WV-3 weights. Runtime includes training and inference. Results are reported as mean $\pm$ standard
  deviation. (Bold: best; Underline: second best)}
  \label{tab:wv3_wv2_comparison}
  \resizebox{\textwidth}{!}{%
    \begin{tabular}{@{}l|cccc|cccc@{}}
      \hline\hline
      \multirow{2}{*}{Method}
        & \multicolumn{4}{c|}{WV-3 (In-Sensor)}
        & \multicolumn{4}{c}{WV-2 (Cross-Sensor)} \\
      \cline{2-9}
        & HQNR $\uparrow$ & $D_\lambda \downarrow$ & $D_s \downarrow$ & Time (s) $\downarrow$
        & HQNR $\uparrow$ & $D_\lambda \downarrow$ & $D_s \downarrow$ & Time (s) $\downarrow$ \\
      \hline
      PSDip \smallgrey{(24'TGRS)} \cite{rui2024variational}
        & $0.9222\pm0.0172$ & $0.0192\pm0.0076$ & $0.0598\pm0.0112$ & $110.079\pm0.022$
        & $0.8935\pm0.0259$ & $0.0412\pm0.0276$ & $0.0680\pm0.0133$ & $110.085\pm0.028$ \\
      ZS-Pan \smallgrey{(24'IF)} \cite{cao2024zero}
        & $0.9454\pm0.0209$ & $0.0252\pm0.0071$ & $0.0302\pm0.0153$ & $3.023\pm0.114$
        & $0.9113\pm0.0331$ & $0.0479\pm0.0271$ & $0.0430\pm0.0126$ & $3.071\pm0.255$ \\
      UCL \smallgrey{(26'IF)} \cite{xiao2025unsupervised}
        & $0.9410\pm0.0131$ & $\underline{0.0103\pm0.0035}$ & $0.0492\pm0.0115$ & $258.640\pm0.355$
        & $0.9221\pm0.0256$ & $\mathbf{0.0222\pm0.0155}$ & $0.0570\pm0.0205$ & $258.562\pm0.475$ \\
      FusionNet \smallgrey{(21'TGRS)} \cite{deng2021fusionnet}
        & $0.9414\pm0.0196$ & $0.0216\pm0.0079$ & $0.0378\pm0.0142$ & $\mathbf{0.009\pm0.001}$
        & $0.8906\pm0.0197$ & $0.0505\pm0.0249$ & $0.0618\pm0.0176$ & $\mathbf{0.009\pm0.001}$ \\
      U2Net \smallgrey{(23'MM)} \cite{peng2023u2net}
        & $0.9507\pm0.0114$ & $0.0174\pm0.0069$ & $0.0325\pm0.0073$ & $0.073\pm0.009$
        & $0.8687\pm0.0842$ & $0.0950\pm0.0826$ & $0.0407\pm0.0100$ & $0.070\pm0.002$ \\
      ERFT$_{\mathrm{FusionNet}}$ \smallgrey{(26'AAAI)} \cite{xin2026training}
        & $0.9478\pm0.0172$ & $0.0190\pm0.0071$ & $0.0338\pm0.0127$ & $\underline{0.059\pm0.003}$
        & $0.9076\pm0.0240$ & $0.0460\pm0.0221$ & $0.0485\pm0.0160$ & $\underline{0.058\pm0.003}$ \\
      ERFT$_{\mathrm{U2Net}}$ \smallgrey{(26'AAAI)} \cite{xin2026training}
        & $0.9586\pm0.0122$ & $0.0202\pm0.0077$ & $\mathbf{0.0216\pm0.0085}$ & $0.198\pm0.007$
        & $0.9221\pm0.0478$ & $0.0564\pm0.0452$ & $\mathbf{0.0230\pm0.0063}$ & $0.202\pm0.019$ \\
      \hline
      \rowcolor{gray!20}Ours$_{\mathrm{FusionNet}}$
        & $\mathbf{0.9661\pm0.0049}$ & $\mathbf{0.0079\pm0.0037}$ & $0.0263\pm0.0042$ & $0.422\pm0.019$
        & $\underline{0.9434\pm0.0252}$ & $\underline{0.0249\pm0.0162}$ & $0.0326\pm0.0133$ & $0.439\pm0.016$ \\
      \rowcolor{gray!20}Ours$_{\mathrm{U2Net}}$
        & $\underline{0.9615\pm0.0115}$ & $0.0133\pm0.0055$ & $\underline{0.0256\pm0.0079}$ & $7.554\pm1.212$
        & $\mathbf{0.9450\pm0.0263}$ & $0.0289\pm0.0174$ & $\underline{0.0270\pm0.0126}$ & $7.797\pm0.968$ \\
      \hline\hline
    \end{tabular}%
  }
\end{table*}

\subsection{Ablation Study}

\paragraph{Effect of parameter selection:}

\Cref{tab:ablation_selection} compares the proposed cross-scoring strategy with fixed parameter selection from the early, middle, and late layers of the pretrained network. The proposed cross-scoring strategy achieves the highest HQNR and lowest spectral distortion on both backbones, demonstrating its effectiveness in identifying beneficial parameters for instance-specific adaptation. Among the above three fixed parameter selection strategies, middle selection performs best on FusionNet \cite{deng2021fusionnet}, whereas late selection achieves the best results on U2Net \cite{peng2023u2net}. These differences indicate that the effectiveness of parameter adaptation depends on both the parameter location and the pretrained network architecture. In contrast, the proposed cross-scoring strategy adaptively selects parameter groups for each test pair, consistently achieving superior overall reconstruction quality across different backbones.

\paragraph{Effect of the consistency losses:}
\Cref{tab:ablation_losses} analyzes the respective contributions of the spatial and spectral losses. Using the spatial loss alone significantly degrades fusion quality, indicating that gradient-based detail injection without spectral constraints may introduce spatial structures that are inconsistent with the observed multispectral information. In contrast, the spectral loss alone better preserves radiometric information but does not sufficiently constrain spatial distortion, potentially leading to performance degradation compared with the pretrained backbone. Combining the two objectives achieves the best overall balance between spectral preservation and spatial detail enhancement. Specifically, the spectral consistency term serves as an observation-based constraint that ensures consistency between the reconstructed image and the observed LRMS image, while the spatial consistency term introduces PAN-guided details to compensate for the spatial information not adequately recovered by the pretrained network. Furthermore, the above \Cref{fig:spa} shows that the proposed spatial fidelity term achieves better reconstruction performance than the two compared spatial fidelity terms.

\subsection{Discussion}

\paragraph{Comparison with Instance-Specific Methods:}
\Cref{tab:wv3_wv2_comparison} compares the proposed framework with recent instance-specific methods under both in-sensor and cross-sensor settings. Methods involving extensive test-time optimization, such as PSDip \cite{rui2024variational}, typically require substantial computational time. The proposed framework achieves the highest HQNR scores on both datasets while maintaining competitive computational efficiency. This can be attributed to the proposed adaptation strategy: cross-scoring evaluates candidate parameter groups using shared gradients without separately optimizing each group, while subsequent adaptation updates only the selected parameters. ERFT \cite{xin2026training} reduces adaptation overhead at the expense of reconstruction quality. The longer runtime of the U2Net-based variant further indicates that adaptation efficiency depends on the backbone architecture and its candidate parameter groups.

\begin{figure}[htbp!]
    \centering
    \includegraphics[width=0.85\linewidth]{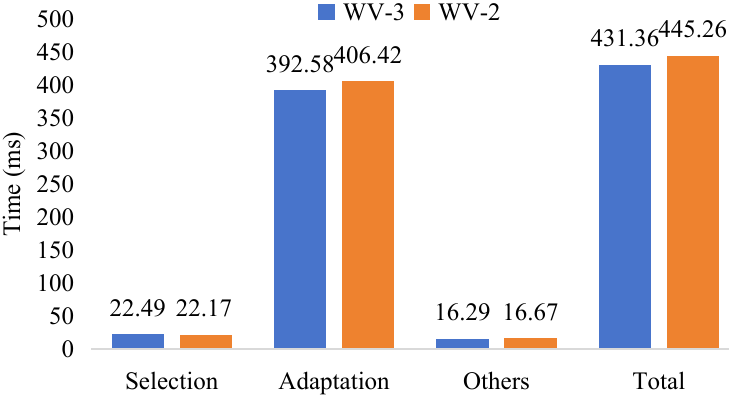}
    \caption{Time analysis.}
    \label{fig:time}
\end{figure}

\begin{figure}[htbp!]
    \centering
    \includegraphics[width=0.8\linewidth]{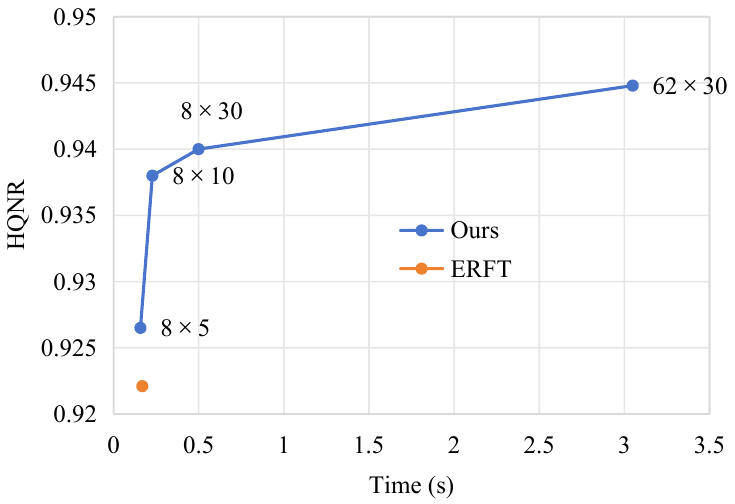}
    \caption{Balance between accuracy and time. In $a \times b$, $a$ denotes the number of training patches and $b$ denotes update steps.}
    \label{fig:balance}
\end{figure}

\paragraph{Time Analysis:} 
In this part, we further analyze the computational time.
\Cref{fig:time} shows that iterative adaptation accounts for the majority of the computational cost. Therefore, acceleration should primarily focus on the parameter update stage. \Cref{fig:balance} illustrates the trade-off between time and accuracy based on U2Net \cite{peng2023u2net}. The proposed method can be accelerated by reducing the number of adaptation iterations and number of training patches. Although this may lead to a slight degradation in reconstruction quality, the proposed method still outperforms ERFT \cite{xin2026training}.

\paragraph{More Experimental Comparisons:} We further provide implementation details, simulated-data experiments, the choice of pretrained networks, and more visual comparisons in the supplementary material.

\paragraph{Limitations:} 
Although our instance-specific adaptation framework has demonstrated its effectiveness across multiple pretrained networks, applying our framework to a new backbone requires manually defining architecture-specific candidate subsets. Besides, the underlying relationship between the pretrained networks and the framework remains unclear and warrants further investigation in future work.

\section{Conclusion}

In this work, we presented an instance-specific adaptation framework consisting of cross-scoring-based parameter selection and test-time adaptation for real-world pansharpening. Experiments on multiple in-sensor and cross-sensor datasets demonstrate that the proposed framework consistently improves the performance of pretrained backbones. Further analyses highlight its advantages over existing instance-specific methods and reveal the trade-off between accuracy and efficiency under different adaptation budgets. Future work will explore the underlying relationship between pretrained networks and the framework.

{
    \small
    \bibliographystyle{ieeenat_fullname}
    \bibliography{main}
}

\clearpage
\setcounter{page}{1}
\maketitlesupplementary

\setcounter{section}{0}

This supplementary material provides implementation details, simulated-data experiments, choice of pretrained networks, and more visual comparisons.

\section{Implementation Details}
\label{sec:supp_impl}

Our framework is implemented and evaluated on a workstation equipped
with a single NVIDIA GeForce RTX 5090 D v2 GPU with 24\,GB of GPU memory.
The workstation runs 64-bit Windows 11 Pro. The implementation uses
Python 3.10.21 and PyTorch 2.11.0 with CUDA 12.8.
Image quality metrics are computed using MATLAB R2025b.

We evaluate FusionNet~\cite{deng2021fusionnet} and
U2Net~\cite{peng2023u2net} on the full-resolution PanCollection
splits~\cite{deng2022machine}. Each split contains 20 pairs of
$512\times512$ PAN and $128\times128$ LRMS images.
For each backbone, we use the QB-pretrained checkpoint on QB and GF-2,
and the WV-3-pretrained checkpoint on WV-3 and WV-2.
Thus, QB and WV-3 are evaluated under the in-sensor setting,
whereas GF-2 and WV-2 are evaluated under the cross-sensor setting.
Direct inference and test-time adaptation start from the same
checkpoint for each backbone and target sensor.

\subsection{Candidate Parameter Groups}
\label{sec:supp_candidate_groups}

A candidate specifies which pretrained weight tensors can be updated.
Candidates within each pool contain the same number of scalar weights,
and all biases are frozen. We define the pools as follows.

\paragraph{FusionNet:}
Each candidate contains either the input convolution
(\texttt{conv1.weight}) or the output convolution
(\texttt{conv3.weight}), together with two distinct residual convolution
tensors. The residual tensors are the weights of \texttt{conv20} and
\texttt{conv21} in each of the four residual blocks, \texttt{res1}
through \texttt{res4}. This gives $2\binom{8}{2}=56$ candidates for
both four-band and eight-band inputs.

\paragraph{U2Net:}
Each candidate contains the final output convolution
(\texttt{to\_hrms.2.weight}) and two groups from
Table~\ref{tab:supp_u2net_groups}. The two convolution tensors in
$G_4$ are selected together. For the eight-band model, all four
groups are eligible, giving $\binom{4}{2}=6$ candidates. For the
four-band model, the input upsampling tensor in $G_1$ contains
$2{,}304$ weights, while each of the other groups contains $9{,}216$.
We use $G_2$, $G_3$, and $G_4$ to form the three equal-budget candidates.

\begin{table}[t]
    \centering
    \footnotesize
    \caption{U2Net groups eligible for selection in addition to the fixed
    inclusion of \texttt{to\_hrms.2.weight} in each candidate.}
    \label{tab:supp_u2net_groups}
    \begin{tabular*}{0.95\columnwidth}
    {@{\extracolsep{\fill}}cl@{}}
        \toprule
        Group & Weight tensor in the checkpoint \\
        \midrule
        $G_1$ & \texttt{upsample.upsamle.0.weight} \\
        $G_2$ & \texttt{to\_hrms.0.weight} \\
        $G_3$ & \texttt{up1.conv.weight} \\
        $G_4$ & \texttt{resblock0.conv0.weight} \\
              & \texttt{resblock0.conv1.weight} \\
        \bottomrule
    \end{tabular*}
\end{table}

For both backbones, the resulting budget for $C$ spectral bands is
\[
    B(C)=2(9{,}216)+32C(3\times3),
\]
which gives $19{,}584$ trainable weights on QB and GF-2 and $20{,}736$
on WV-3 and WV-2.

\subsection{Parameter Selection and Updating}
\label{sec:supp_full_selection}

\paragraph{Cross scoring:}

For the full-resolution experiments in the main paper, each
observation pair contains a $512\times512$ PAN image and a
$128\times128$ LRMS image.
We extract one $32\times32$ LRMS patch and its aligned
$128\times128$ PAN patch from the center of each quadrant. After MTF filtering and downsampling, the original LRMS patch serves as the reference for the reduced-resolution reconstruction task. Diagonally opposite patches are grouped into one class. We compute the reconstruction MSE over the central $24\times24$ region and obtain one gradient per class at the pretrained weights.

For each candidate, we perform a virtual Adam step using the gradient from one class and estimate the MSE reduction on the other class through a first-order approximation. We repeat this process with the two classes exchanged and average the resulting scores. The virtual step uses a learning rate and coupled weight decay of $10^{-5}$, with $\epsilon=10^{-8}$ and zero-initialized moments. The two class gradients are shared across all candidates. We select the highest-scoring candidate for each test pair and perform the subsequent adaptation starting from the original pretrained weights.

\begin{table}[t]
    \centering
    \footnotesize
    \caption{Full-resolution update settings. The step count gives
    the update cap for runs with held-out checkpoint selection.}
    \label{tab:supp_full_update_settings}
    \begin{tabular*}{0.95\columnwidth}
    {@{\extracolsep{\fill}}llccc@{}}
        \toprule
        Target & Backbone & $\lambda$ & Adam LR & Steps $N$ \\
        \midrule
        WV-3 / WV-2 & Both      & $0.03$   & $4\times10^{-4}$   & 30 \\
        QB          & FusionNet & $0.001$  & $6\times10^{-4}$   & 20 \\
        QB          & U2Net     & $0.001$  & $6\times10^{-4}$   & 35 \\
        GF-2        & FusionNet & $0.03$   & $4\times10^{-4}$   & 30 \\
        GF-2        & U2Net     & $0.0003$ & $1.2\times10^{-3}$ & 45 \\
        \bottomrule
    \end{tabular*}
\end{table}

\begin{figure*}[t]
    \centering
    \includegraphics[width=1\linewidth]{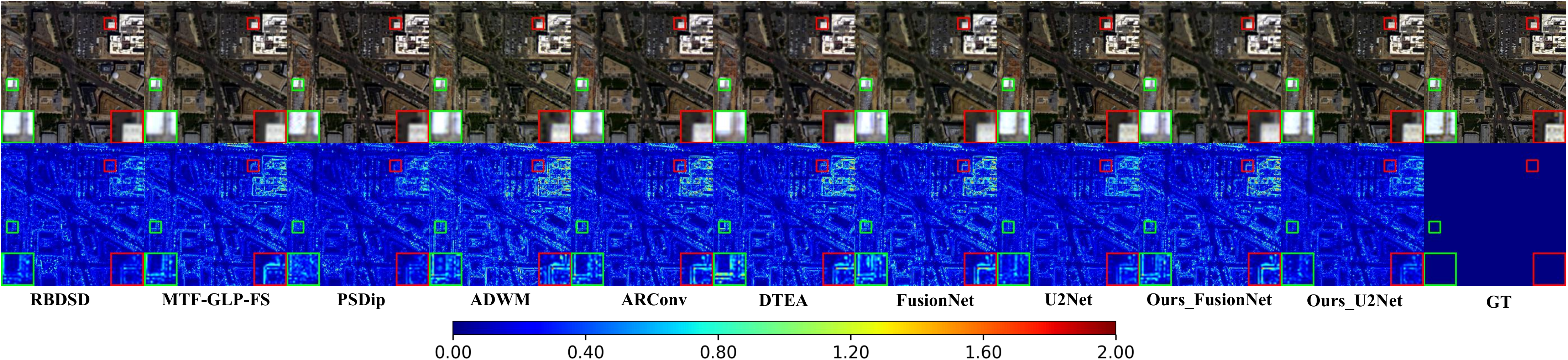}
    \caption{The visual comparisons on the WV-2 dataset on reduced scale. The upper row shows the predicted HRMS images of different methods, and the bottom row displays the corresponding residual maps.  }
    \label{fig:supp1}
\end{figure*}

\begin{table*}[t]
  \centering
  \footnotesize
  \setlength{\tabcolsep}{1.0mm}
  \renewcommand{\arraystretch}{1.2}
  \caption{Reduced-resolution quantitative comparison on the WV-2 dataset.
  Results are reported as mean $\pm$ sample standard deviation
  over 20 image pairs. (Bold: best; Underline: second best)}
  \label{tab:wv2_reduced}
  \resizebox{\textwidth}{!}{%
    \begin{tabular}{@{}l|cccccc@{}}
      \hline\hline
      Method
        & PSNR $\uparrow$
        & SSIM $\uparrow$
        & SAM $\downarrow$
        & ERGAS $\downarrow$
        & SCC $\uparrow$
        & Q8 $\uparrow$ \\
      \hline
      RBDSD \smallgrey{(19'TGRS)} \cite{vivone2019robust}
        & $30.4044\pm1.4992$
        & $0.8365\pm0.0282$
        & $6.2846\pm0.9200$
        & $4.2758\pm0.7420$
        & $0.9089\pm0.0204$
        & $0.8497\pm0.0800$ \\
      MTF-GLP-FS \smallgrey{(18'TIP)} \cite{vivone2018full}
        & $29.4491\pm1.5268$
        & $0.8062\pm0.0357$
        & $6.4338\pm0.8414$
        & $4.7346\pm0.7219$
        & $0.8876\pm0.0254$
        & $0.8191\pm0.0804$ \\
      PSDip \smallgrey{(24'TGRS)} \cite{rui2024variational}
        & $\underline{30.9594\pm1.6543}$
        & $0.8575\pm0.0260$
        & $5.5738\pm0.6284$
        & $\underline{3.9180\pm0.5265}$
        & $0.9286\pm0.0076$
        & $\underline{0.8618\pm0.0835}$ \\
      ADWM \smallgrey{(25'CVPR)} \cite{huang2025general}
        & $28.0522\pm1.6646$
        & $0.7461\pm0.0467$
        & $6.3642\pm0.5456$
        & $5.4875\pm0.6386$
        & $0.8622\pm0.0213$
        & $0.7696\pm0.0938$ \\
      U2Net \smallgrey{(23'MM)} \cite{peng2023u2net}
        & $30.4962\pm1.8092$
        & $\underline{0.8589\pm0.0239}$
        & $\underline{5.2586\pm0.4992}$
        & $4.0766\pm0.3832$
        & $\mathbf{0.9333\pm0.0071}$
        & $0.8470\pm0.0846$ \\
      DTEA \smallgrey{(26'TIP)} \cite{xu2026deta}
        & $29.5017\pm2.1061$
        & $0.8101\pm0.0428$
        & $5.7525\pm0.5679$
        & $4.6171\pm0.3739$
        & $0.9016\pm0.0125$
        & $0.8249\pm0.0768$ \\
      FusionNet \smallgrey{(21'TGRS)} \cite{deng2021fusionnet}
        & $28.5405\pm2.0037$
        & $0.7866\pm0.0431$
        & $6.4118\pm0.6231$
        & $5.1429\pm0.4562$
        & $0.8788\pm0.0142$
        & $0.7928\pm0.0757$ \\
      ARConv \smallgrey{(25'CVPR)} \cite{Wang2025CVPR}
        & $29.8534\pm2.0310$
        & $0.8240\pm0.0374$
        & $5.4516\pm0.5929$
        & $4.4176\pm0.4099$
        & $0.9123\pm0.0093$
        & $0.8357\pm0.0775$ \\
      \rowcolor{gray!20}
      Ours$_{\mathrm{FusionNet}}$
        & $28.8155\pm2.0212$
        & $0.7931\pm0.0428$
        & $6.1274\pm0.6269$
        & $4.9821\pm0.4248$
        & $0.8876\pm0.0121$
        & $0.8005\pm0.0772$ \\
      \rowcolor{gray!20}
      Ours$_{\mathrm{U2Net}}$
        & $\mathbf{31.0316\pm1.7817}$
        & $\mathbf{0.8653\pm0.0243}$
        & $\mathbf{5.2049\pm0.5427}$
        & $\mathbf{3.8605\pm0.4179}$
        & $\underline{0.9331\pm0.0082}$
        & $\mathbf{0.8647\pm0.0823}$ \\
      \hline\hline
    \end{tabular}%
  }
\end{table*}

\paragraph{Full-resolution updates:}
We divide each image pair into 64 aligned patches with $64\times64$
PAN cores and $16\times16$ LRMS cores. Each update uses 62 patches.
For configurations with checkpoint selection, the remaining two
patches serve as the validation set. We update only the selected
weights using Adam with weight decay $10^{-5}$ and the loss objective
$\mathcal{L}_{\mathrm{spe}}+
\lambda\mathcal{L}_{\mathrm{spa}}$ (as shown in \cref{eq:joint_adaptation_loss}).
For runs with held-out checkpoint selection, we evaluate this
objective at step zero and every five updates, stop after three
consecutive checks without improvement, and restore the checkpoint
with the lowest validation loss. The trade-off parameter $\lambda$,
Adam learning rate (LR), and number of update steps $N$ are
hyperparameters that require tuning. Their detailed settings are
provided in \Cref{tab:supp_full_update_settings}.

\section{Simulated-Data Experiment}
\label{sec:supp_simulated}
Our work primarily focuses on real-world pansharpening.
Although ground-truth HRMS images are unavailable at the native
PAN resolution, reduced-resolution simulations allow for
full-reference evaluation of pansharpening methods.
Therefore, we complement the full-resolution experiments with
simulated-data experiments using the released reduced-resolution
WV-2 dataset from PanCollection~\cite{deng2022machine}. This dataset contains 20 image pairs, each consisting of a
$256\times256$ PAN image, a $64\times64\times8$ LRMS image,
and a $256\times256\times8$ reference HRMS image.
We evaluate the reconstruction performance using six
widely adopted quality metrics: PSNR, SSIM, SAM, ERGAS,
SCC, and Q8.


\paragraph{Data simulation:}
The simulated data are constructed following Wald's protocol \cite{wald1997protocol}.
Given a spatially aligned source pair consisting of an MS image
$\mathbf{Y} \in \mathbb{R}^{H \times W \times C}$ and a PAN image
$\mathbf{P} \in \mathbb{R}^{rH \times rW}$, both observations are
spatially degraded by a factor of $r=4$.
The resulting inputs and reference image are expressed as
\begin{equation}
\begin{aligned}
    \mathbf{Y}_{\mathrm{sim}}
        &= \mathtt{MTF}(\mathbf{Y})\downarrow_r, \\
    \mathbf{P}_{\mathrm{sim}}
        &= \mathtt{MTF}_{\mathrm{PAN}}(\mathbf{P})\downarrow_r, \\
    \mathbf{X}_{\mathrm{ref}}
        &= \mathbf{Y},
\end{aligned}
\label{eq:supp_simulation}
\end{equation}
where $\mathtt{MTF}$ and
$\mathtt{MTF}_{\mathrm{PAN}}$ denote the respective low-pass
filtering operators for multispectral and PAN images, respectively, and $\downarrow_r$ is the down-sampling operation with scale ratio $r$.
Here, $\mathbf{Y}_{\mathrm{sim}}$ and $\mathbf{P}_{\mathrm{sim}}$
serve as the inputs, while $\mathbf{X}_{\mathrm{ref}}$
serves as the reference HRMS image for reduced-resolution evaluation.
The MS degradation incorporates filtering based on the sensor
modulation transfer function.
This procedure preserves the spatial resolution ratio between
the simulated PAN and LRMS inputs, while retaining the original
MS image as the reference at the simulated target resolution.

\begin{figure*}[t]
    \centering
    \includegraphics[width=1\linewidth]{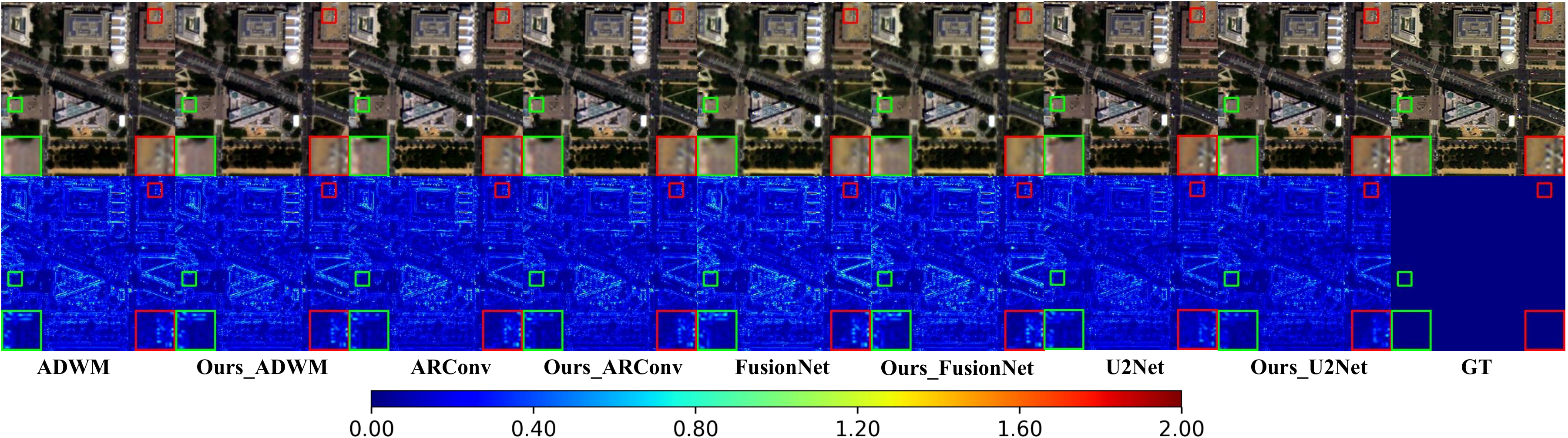}
    \caption{The visual comparisons with different pretrained networks on the WV-2 dataset. The upper row shows the predicted HRMS images of different methods, and the bottom row displays the corresponding residual maps.  }
    \label{fig:supp2}
\end{figure*}

\begin{table*}[htbp!]
  \centering
  \footnotesize
  \setlength{\tabcolsep}{1.0mm}
  \renewcommand{\arraystretch}{1.2}
  \caption{Reduced-resolution cross-sensor comparison on WV-2 dataset.
  Results are reported as mean $\pm$ sample standard deviation
  over 20 image pairs. (Bold: best; Underline: second best)}
  \label{tab:wv2_reduced_backbones}
  \resizebox{\textwidth}{!}{%
    \begin{tabular}{@{}l|cccccc@{}}
      \hline\hline
      Method
        & PSNR $\uparrow$
        & SSIM $\uparrow$
        & SAM $\downarrow$
        & ERGAS $\downarrow$
        & SCC $\uparrow$
        & Q8 $\uparrow$ \\
      \hline
      ADWM
        & $29.6009\pm1.9074$
        & $0.81595\pm0.03646$
        & $5.8006\pm0.6917$
        & $4.5894\pm0.4466$
        & $0.91267\pm0.00954$
        & $0.82165\pm0.07905$ \\
      \rowcolor{gray!20}
      Ours$_{\mathrm{ADWM}}$
        & $29.7683\pm1.9628$
        & $0.82162\pm0.03596$
        & $5.6955\pm0.6496$
        & $4.5007\pm0.4354$
        & $0.91516\pm0.00873$
        & $0.82823\pm0.07839$ \\
      \hline
      ARConv
        & $29.8534\pm2.0310$
        & $0.82397\pm0.03736$
        & $5.4516\pm0.5929$
        & $4.4176\pm0.4099$
        & $0.91233\pm0.00931$
        & $0.83569\pm0.07745$ \\
      \rowcolor{gray!20}
      Ours$_{\mathrm{ARConv}}$
        & $29.9030\pm2.0288$
        & $0.82485\pm0.03715$
        & $5.4411\pm0.5971$
        & $4.3962\pm0.4100$
        & $0.91226\pm0.00937$
        & $0.83781\pm0.07747$ \\
      \hline
      FusionNet
        & $28.5405\pm2.0037$
        & $0.78660\pm0.04308$
        & $6.4118\pm0.6231$
        & $5.1429\pm0.4562$
        & $0.87878\pm0.01420$
        & $0.79285\pm0.07571$ \\
      \rowcolor{gray!20}
      Ours$_{\mathrm{FusionNet}}$
        & $28.8155\pm2.0212$
        & $0.79306\pm0.04283$
        & $6.1274\pm0.6269$
        & $4.9821\pm0.4248$
        & $0.88758\pm0.01208$
        & $0.80052\pm0.07715$ \\
      \hline
      U2Net
        & $\underline{30.4962\pm1.8092}$
        & $\underline{0.85891\pm0.02388}$
        & $\underline{5.2586\pm0.4992}$
        & $\underline{4.0766\pm0.3832}$
        & $\mathbf{0.93326\pm0.00713}$
        & $\underline{0.84696\pm0.08457}$ \\
      \rowcolor{gray!20}
      Ours$_{\mathrm{U2Net}}$
        & $\mathbf{31.0316\pm1.7817}$
        & $\mathbf{0.86528\pm0.02432}$
        & $\mathbf{5.2049\pm0.5427}$
        & $\mathbf{3.8605\pm0.4179}$
        & $\underline{0.93314\pm0.00816}$
        & $\mathbf{0.86467\pm0.08234}$ \\
      \hline\hline
    \end{tabular}%
  }
\end{table*}

\paragraph{Experimental results:}
To further evaluate the effectiveness of the proposed framework,
we conduct reduced-resolution cross-sensor experiments on WV-2
using pretrained networks trained on WV-3.
As shown in \Cref{tab:wv2_reduced}, the proposed framework
consistently improves the performance of both FusionNet and U2Net
across most evaluation metrics.
Specifically, Ours$_{\mathrm{U2Net}}$ improves the PSNR of U2Net and achieves the best results in five of the six metrics, while maintaining comparable SCC performance.
Similarly, Ours$_{\mathrm{FusionNet}}$ improves the PSNR by 0.28 dB,
with consistent improvements across all six metrics.
Compared with other state-of-the-art methods,
Ours$_{\mathrm{U2Net}}$ achieves favorable reconstruction
performance, demonstrating the effectiveness of the proposed
framework under cross-sensor conditions.
\Cref{fig:supp1} presents the corresponding visual comparisons
and residual maps, providing a qualitative assessment of the
reconstruction results. As shown in \Cref{fig:supp1}, the proposed method preserves more spatial details compared to other methods.

\section{ Choice of Pretrained Networks}
\label{sec:supp_pretrain}

In this section, we investigate the influence of pretrained networks on the proposed framework. U2Net \cite{peng2023u2net} and FusionNet \cite{deng2021fusionnet} are selected as representative pretrained networks in the main text. To further evaluate the generality of the proposed framework, we extend the experiments to two additional networks, ADWM \cite{huang2025general} and ARConv \cite{Wang2025CVPR}. All networks are pretrained on the WV-3 dataset and evaluated on the WV-2 dataset under the reduced-resolution setting to assess their cross-sensor generalization ability.

As shown in \Cref{tab:wv2_reduced_backbones}, the proposed framework consistently improves the reconstruction performance of different pretrained
networks. Specifically, the PSNR improvements over ADWM, ARConv, FusionNet, and U2Net are 0.17 dB, 0.05 dB, 0.28 dB, and 0.54 dB, respectively.
Improvements are also observed in SSIM, SAM, ERGAS, and Q8 across all four networks, while SCC remains comparable with minor variations. Notably, the
proposed framework achieves the best performance when combined with U2Net, attaining a PSNR of 31.03 dB. The visual comparisons in \Cref{fig:supp2} further illustrate the reconstruction results and corresponding residual
maps obtained with different pretrained networks. These results demonstrate that the proposed framework is applicable to different pretrained networks and can further enhance their reconstruction performance under cross-sensor conditions.

\end{document}